# Grasp Stability Assessment Through Attention-Guided Cross-Modality Fusion and Transfer Learning

Zhuangzhuang Zhang[1], Zhenning Zhou[1], Haili Wang[1], Zhinan Zhang[1], Huang Huang[2] and Qixin Cao[1]

*Abstract*—Extensive research has been conducted on assessing grasp stability, a crucial prerequisite for achieving optimal grasping strategies, including the minimum force grasping policy. However, existing works employ basic feature-level fusion techniques to combine visual and tactile modalities, resulting in the inadequate utilization of complementary information and the inability to model interactions between unimodal features. This work proposes an attention-guided cross-modality fusion architecture to comprehensively integrate visual and tactile features. This model mainly comprises convolutional neural networks (CNNs), self-attention, and cross-attention mechanisms. In addition, most existing methods collect datasets from real-world systems, which is time-consuming and high-cost, and the datasets collected are comparatively limited in size. This work establishes a robotic grasping system through physics simulation to collect a multimodal dataset. To address the sim-to-real transfer gap, we propose a migration strategy encompassing domain randomization and domain adaptation techniques. The experimental results demonstrate that the proposed fusion framework achieves markedly enhanced prediction performance (approximately 10%) compared to other baselines. Moreover, our findings suggest that the trained model can be reliably transferred to real robotic systems, indicating its potential to address real-world challenges.

## I. INTRODUCTION

Before grasping an object, humans effortlessly integrate the senses of vision and touch to assess the stability of the grasp. Visual feedback provides information regarding the geometric properties of the object's surface, while tactile feedback establishes precise and intuitive contact conditions between the hand and the object. Thus, these two modalities are concurrent and complementary. However, the existing robotic grasping methodologies typically use a fixed gripping force. As a result, the robot primarily relies on open-loop grasping and cannot actively modify its pose and gripping force, thereby limiting the stability and security of the grasp. Additionally, it is critical to equip robots with the capability to delicately and minimally grasp objects, much like humans, as this can substantially enhance robots' intelligence in handling the complexities encountered in unstructured environments [1]. The grasp stability assessment, which serves as a critical prerequisite for enabling intelligent grasping, remains an open and challenging research issue [2], [3].

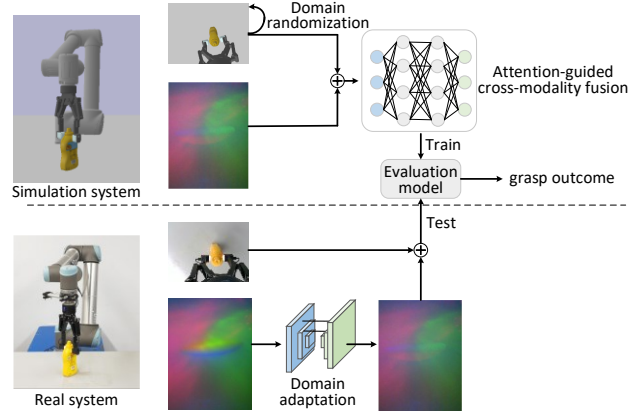

**Fig. 1:** The framework diagram of the proposed grasp stability assessment method: the prediction network is trained with synthetic visual and tactile images and then successfully deployed on a real robot using the proposed migration strategy.

To date, several typical studies on the assessment of grasp stability have focused on multimodal fusion networks (MMFNs) that utilize both visual and tactile modalities. Calandra *et al.* [2] proposed a multimodal sensing framework for predicting grasp stability using tactile and visual inputs. Their experimental results indicate that the visual-tactile model significantly enhances grasping performance. They further introduced a regrasping policy based on grasp stability evaluation using raw visual-tactile data. The learned model enables the robot to grasp objects with minimal gripping force, reducing the chance of object damage [3]. Li *et al.* [4] introduced an architecture constructed from CNN and Recurrent Neural Network (RNN) to classify a grasp as stable or not. Cui *et al.* [5] proposed a 3D CNN-based fusion perception network to evaluate the grasp stability of deformable objects. They also introduced an MMFN that utilizes the self-attention mechanism [6]. In a recent study, Kanitkar *et al.* [7] introduced a multimodal dataset that includes tactile and visual data to explore grasp outcomes at specific holding poses. However, the visual-tactile fusion-based grasp stability evaluation methods discussed above still exhibit certain limitations.

Firstly, these studies employed simple feature-level fusion techniques (e.g., concatenation of the unimodal features from the final layer) to train multimodal prediction networks. Even in work [6], only one single-layer self-attention module is utilized. This has led to insufficient utilization of complementary information and a failure to model interactions between unimodal features. In recent years, transformers have been demonstrated to perform well across various tasks, such as natural language processing (NLP) [8],

This work was supported by China's National Key Research and Development Program under Grant 2018AAA0102700. (corresponding author: Qixin Cao qxcao@sjtu.edu.cn)

[1]Zhuangzhuang Zhang, Zhenning Zhou, Haili Wang, Zhinan Zhang and Qixin Cao are with the School of Mechanical Engineering, Shanghai Jiao Tong University, Shanghai 200240, P.R. China zhangzzsjtu@gmail.com, {zhenning_zhou, sjtu_wanghaili, zhinanz, qxcao}@sjtu.edu.cn

[2]Huang Huang is with the Beijing Institute of Control Engineering, Beijing 100191, P.R. China hhuang33@163.com

visuo-tactile manipulation [35], motion forecasting [36], as well as processing multimodal data, such as images, audio, and video [9]. Inspired by these observations, we propose an attention-guided visual-tactile cross-modality fusion method for delicate robotic grasping tasks. Specifically, this architecture utilizes a self-attention-based module to enhance unimodal information, a cross-attention-based module to model the interactions between unimodal features, and a co-attention module to aggregate and enhance visual-tactile features.

Secondly, these methods involve the collection of datasets from real-world systems, which is a time-consuming and expensive process, and the resulting datasets are often limited in size. A large-scale and reasonable dataset is a primary prerequisite for data-driven methods. To accelerate the dataset generation process, applying physics simulation provides an appealing avenue. Therefore, in this paper, we set up a robotic grasping system in the physics simulator PyBullet and implement a visual-tactile multimodal dataset collection policy. We then propose a migration strategy that consists of domain randomization and domain adaptation techniques to bridge the sim-to-real transfer gap. The framework diagram of the proposed grasp stability assessment method is shown in Fig. 1. The contributions of this paper are summarised as follows:

(1) An end-to-end attention-guided cross-modality fusion architecture is proposed to assess the grasp stability.

(2) A migration strategy that consists of domain randomization and domain adaptation techniques is proposed to bridge the sim-to-real transfer gap.

(3) Extensive validation experiments are conducted in both real and simulation systems, and the results prove that the proposed model outperforms other baselines and can be reliably transferred to the real robotic system.

The remaining part of this paper is structured as follows. Section II reviews the related work of grasp stability evaluation and sim-to-real transfer. In Section III, the cross-modality fusion architecture is described. In Section IV, the dataset generation and migration strategies are presented. In Section V, extensive validation experiments are conducted in simulated and real systems, and the experimental results are discussed. Finally, Section VI is the conclusion of this paper and future work.

## II. RELATED WORK

### A. Grasp Stability Evaluation

Grasp stability evaluation has been extensively researched as a crucial prerequisite for optimizing grasping strategies. Bekiroglu et al. [10] introduced a probabilistic learning framework that utilizes machine learning techniques and tactile data acquired from pressure-sensitive tactile sensors to evaluate grasp stability. Kwiatkowski et al. [11] utilized CNNs to evaluate grasp stability by combining tactile signals and proprioceptive information. Veiga et al. [12] employed tactile data to predict slip events and modulate contact forces accordingly in anticipation of slip occurrences. Nevertheless, these techniques typically use electronic tactile sensors (ETSs) that offer limited tactile information, impeding robotic tactile sensing performance advancement.

Compared to ETSs, vision-based tactile sensors (VBTSs), such as GelSight-style sensors, offer notable benefits in high-resolution, robustness, and integration of visual-tactile data. Kolamuri et al. [13] employed GelSight sensors to detect the rotational failure of grasp and presented a regrasping strategy to enhance grasp stability. Si et al. [14] developed a CNN-LSTM model that uses a sequence of tactile images to predict grasp outcomes. However, these research frameworks do not integrate visual modality, the concurrent and synergistic integration of visual-tactile data during the initial grasping stage is critical for achieving optimal grasping results. Calandra et al. [2] showed that including tactile signals in a multimodal perception framework significantly improves grasping performance. Cui et al. [5] utilized a 3D CNN-based visual-tactile fusion network to evaluate the grasp state of deformable objects. Kanitkar et al. [7] presented a multimodal dataset consisting of visual-tactile information to investigate the impact of varied holding poses on grasp stability. Nevertheless, the adoption of simplistic feature-level fusion approaches in these studies resulted in a restricted exploitation of complementary information and an inability to effectively capture the interplay among unimodal features. In contrast to these previous studies, we propose an attention-guided cross-modality fusion architecture to enhance unimodal information and model the interactions between unimodal features.

### B. Sim-to-Real Transfer

While generating datasets in simulation is highly efficient, the distributional shift between real and simulation data may lead to migration failure, also known as the sim-to-real gap. By bridging the distribution gap between simulation and the real world, transfer learning can enable the control strategies learned in the simulation to be effectively applied to a real robot. Some works investigated the effectiveness of domain randomization techniques in transferring a model trained on simulated RGB images to real-world images [15], [16]. By introducing randomization in the rendering process within the simulator, these studies successfully reduced the distribution gap between simulated and real-world data and enabled the successful deployment of the trained model on real hardware. Simulating the VBTS is challenging compared to the visual modality because an ideal high-resolution tactile simulator needs to model not only realistic optical properties but also accurate contact dynamics.

Gomes et al. [17] utilized the Gazebo built-in camera to capture the depth map of the contact area and generated the RGB image using Phong's model. Agarwal et al. [18] employed the bidirectional path-tracing algorithm to generate more realistic synthetic images, but this method requires a significant amount of computation. Si et al. [19] proposed Taxim, an example-based method for simulating GelSight sensors that involves optical and marker motion field simulation. Wang et al. [20] proposed TACTO, a simulation framework for simulating VBTSs such as DIGIT [21] and OmniTact [22]. Although the studies mentioned above have shown impressive results, the gap between synthetic tactile images and real images remains due to the challenges in modeling optical properties and contact dynamics. Chen et al. [23] utilized CycleGAN [24] to train unpaired data. Nevertheless, the physical properties of the tactile sensor are neglected. Lin et al. [25] employed an image-to-image translation GAN [26] to accomplish the sim-to-real transfer. Nonetheless, they solely evaluated the zero-shot performance in basic scenarios such as edge-following and surface-following. In this paper, DIGIT [21] and TACTO [20] are employed as tools for capturing real and simulated tactile images, and a migration strategy consisting of domain randomization and domain adaptation techniques is proposed to bridge the sim-to-real transfer gap for delicate grasping tasks.

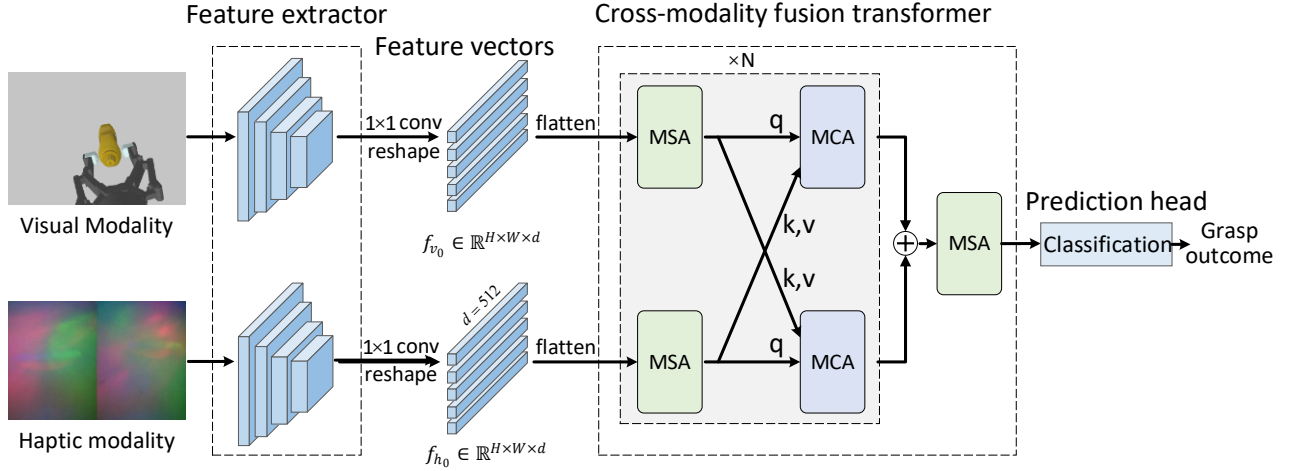

Fig. 2: Dual-stream fusion network architecture for grasp stability prediction.

## III. CROSS-MODALITY FUSION ARCHITECTURE

### A. Network Structure

The proposed dual-stream visual-tactile fusion network in this paper consists of feature extraction, cross-modality fusion transformer, and prediction head, as shown in Fig. 2.

**Feature extraction** The visual modality is an RGB image $\mathcal{I}_v \in \mathbb{R}^{3 \times 480 \times 640}$ captured by a camera mounted on the robot's hand, whereas the tactile modality corresponds to a spliced image $\mathcal{I}_h \in \mathbb{R}^{3 \times 320 \times 480}$ generated from two high-resolution tactile sensors mounted on the gripper. It is noteworthy that this paper investigates the grasp stability assessment. For this purpose, the images of both modalities are captured simultaneously after the gripper closure and prior to object lifting. In contrast, for slip detection purposes, sequential images during the lifting process are required.

We first take ResNet-50 [27] as the backbone for each stream to extract deep features. Specifically, the final outputs of ResNet-50 are obtained by discarding the last stage and using the outputs of the fourth stage instead. Following that, the channel dimension is reduced by a 1×1 convolution to obtain two feature maps with lower dimensions $(\mathbf{X}_{v_0}, \mathbf{X}_{h_0}) \in \mathbb{R}^{H \times W \times d}$. The value of $d$ is set to 512 in our implementation. Finally, the image features are flattened along their spatial dimension into one-dimensional features $(\mathbf{X}_{v_1}, \mathbf{X}_{h_1}) \in \mathbb{R}^{HW \times d}$, which is then utilized as input to the cross-modality fusion transformer.

**Transformer module** First, we employ two multi-head self-attention (MSA) modules to integrate the global interactions and enhance feature representation in the same domain. Then, two multi-head cross-attention modules (MCA) are devised to further integrate global interactions between visual-tactile domains. In this way, a fusion layer is created by combining two MSA modules and two MCA modules. This fusion layer repeats four times in the experiment. The features of the two modalities are then concatenated and fed into an MSA module (also known as the co-attention mechanism) to aggregate the global context, and the final output is a 512-length feature vector.

Given an input sequence $\mathbf{X} \in \mathbb{R}^{HW \times d}$, it will pass through three projection matrices $\mathbf{W}^Q \in \mathbb{R}^{d \times d_k}$, $\mathbf{W}^K \in \mathbb{R}^{d \times d_k}$, and $\mathbf{W}^V \in \mathbb{R}^{d \times d_v}$ to produce three embeddings $\mathbf{Q}$ (Query), $\mathbf{K}$ (Key), and $\mathbf{V}$ (Value):

$$\{\mathbf{Q}, \mathbf{K}, \mathbf{V}\} = \{\mathbf{X}\mathbf{W}^Q, \mathbf{X}\mathbf{W}^K, \mathbf{X}\mathbf{W}^V\} \quad (1)$$

Then the self-attention mechanism is defined as:

$$\mathbf{X} \leftarrow \text{SA}(\mathbf{Q}, \mathbf{K}, \mathbf{V}) = \text{softmax}\left(\frac{\mathbf{Q}\mathbf{K}^T}{\sqrt{d_k}}\right)\mathbf{V} \quad (2)$$

Multiple self-attention sub-layers can be stacked in parallel to consider diverse attention distributions. Thus the structure of multi-head self-attention (MHSA) is defined as:

$$\text{MHSA}(\mathbf{Q}, \mathbf{K}, \mathbf{V}) = \text{Concat}(\mathbf{X}_1, \cdots, \mathbf{X}_{n_h})\mathbf{W}^O \quad (3)$$

where $\mathbf{W}^O \in \mathbb{R}^{n_h d_v \times d}$ is a parameter matrix. In the experiment, we set $n_h = 8$, $d = 512$, and $d_k = d_v = d/n_h = 64$.

In addition, each module is followed by a two-layer feed-forward network (FFN) to enhance the fitting ability of the network,

$$\text{FFN}(\mathbf{X}) = \max(\mathbf{0}, \mathbf{X}\mathbf{W}_1 + \mathbf{b}_1)\mathbf{W}_2 + \mathbf{b}_2 \quad (4)$$

Both FFNs and attention modules employ residual connection. We also apply the positional embedding to both MSA and MCA modules because the attention mechanism cannot distinguish positional information of the input feature sequence. Following [8], we adopt sin and cos functions to encode the positional information $\mathbf{P}$ of the input sequence. Thus, the MSA can be formulated as

$$\mathbf{X} \leftarrow \mathbf{X} + \text{MHSA}(\mathbf{X} + \mathbf{P}, \mathbf{X} + \mathbf{P}, \mathbf{X}) \quad (5)$$

And the MCA can be expressed as

$$X_v \leftarrow X_v + \text{MHSA}(X_v + P_v, X_h + P_h, X_h) \quad (6)$$

$$X_h \leftarrow X_h + \text{MHSA}(X_h + P_h, X_v + P_v, X_v) \quad (7)$$

where $X_v$ and $X_h$ denote the feature sequences of the visual and tactile channels, respectively.

**Prediction head** The prediction head is a classification module consisting of a two-layer FFN structure and a loss function. This module takes 512 feature vectors as input and outputs binary classification results. In addition, a ReLU activation function is employed between the two FFN layers.

### B. Training

In this work, we adopt the standard binary cross-entropy loss to measure the loss of classification,

$$\mathcal{L}_{BCE} = -\sum_i [y_i \log(p_i) + (1 - y_i)\log(1 - p_i)] \quad (8)$$

where $y_i$ represents the ground-truth label, $y_i = 1$ denotes successful grasp, and $p_i$ is the probability of belonging to the successful grasp predicted by the learned model.

We resize tactile and visual images to $256 \times 256 \times 3$ and randomly sample $224 \times 224 \times 3$ crops for data augmentation. The ResNet-50 pretrained by ImageNet [32] is employed. We train 20 epochs on the full network using Adam optimizer with a learning rate of $1 \times 10^{-4}$ and batch size of 32. The experiments are implemented on Ubuntu 18.04 with one NVIDIA GTX1080Ti GPU and a 2.10 GHz Intel Xeon E5-2620 CPU.

## IV. DATASET GENERATION AND MIGRATION STRATEGIES

Collecting large-scale datasets in the real world is time-consuming and laborious. Therefore, robot simulation plays a crucial role in data-driven manipulation tasks. In this paper, we set up a simulation environment for robotic grasping to generate a large-scale visual-tactile dataset and successfully transfer the learned policy to the real world with the proposed migration strategy.

### A. Experimental Conditions

To collect a large and reasonable multimodal dataset, a simulation system for robotic grasping is built. The physical entities in the real-world setting include a UR10 robot, a Robotiq gripper, a RealSense SR305 camera, two high-resolution tactile sensors, and an object that is intended to be grasped. The DIGIT [21] is selected as the tactile sensing hardware for real-world implementation due to its seamless integration with the gripper and user-friendly operation. Simultaneously, the TACTO [20] replicates DIGIT in the simulation environment. Moreover, OpenGL integrated with PyBullet is utilized to render RGB images. The simulated hardware maintains identical CAD dimensions to those of the physical environment and is loaded via URDF. The communication between the two environments is established through the Robot Operating System (ROS).

### B. Visual-Tactile Dataset

This multimodal dataset $\mathcal{D}$ contains tactile images and rendered RGB images. The object set includes 50 objects from the YCB dataset [29]. These objects are divided into a training set of 45 objects and a test set of 5. The overall process framework for visual-tactile dataset generation is shown in Fig. 3. We first sample 400 force closure grasp candidates for each object using the antipodal grasp sampling method [30]. The sampled grasp configurations are represented in the object's reference frame. Then the MeshPy library calculates the stable poses for each object on a table surface and makes corrections in the PyBullet gravity environment, discarding unstable entries. Twenty collision-free grasps are subsequently computed for each stable pose on a planar work surface. Finally, the resulting grasps corresponding to each stable pose are stored in the HDF5 file.

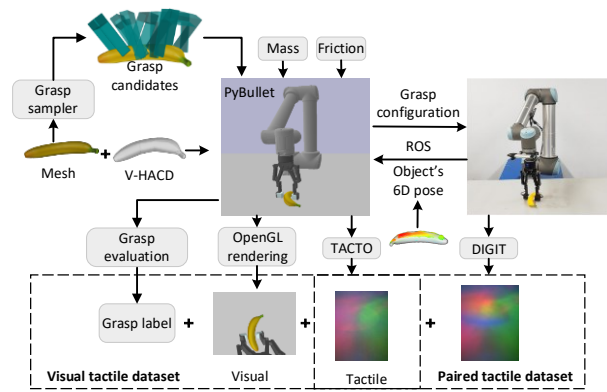

**Fig. 3:** The data collection procedure for visual-tactile and paired tactile datasets.

To simplify the problem, we assume that the grasping is a quasi-static process with a Coulomb friction model, and the friction coefficient is set to $\mu = 0.6$ for objects and gripper in grasping trials. In addition, the mass of the object is taken from the dataset. When performing grasping trials, each object will be loaded with all stored stable poses in order, and each stable pose comes with 20 grasp configurations. The robot will perform 30 grasping trials for each grasp configuration, starting with a gripping force of 5 N and rising at an interval of 1 N. After closing the gripper, images from two tactile sensors are recorded. To initially match the distribution of synthetic images to real readings, a 2D Gaussian filter $G(x, y)$ is first applied,

$$\mathcal{I} \leftarrow \mathcal{I} * \frac{1}{\sqrt{2\pi\sigma^2}} e^{-\frac{x^2+y^2}{2\sigma^2}} \quad (9)$$

where $*$ stands for the convolution operation. Then the synthetic difference image is added to the real background image to form the training input,

$$\mathcal{I} \leftarrow \mathcal{I} - \mathcal{I}_{\text{sim,bg}} + \mathcal{I}_{\text{real,bg}} \quad (10)$$

RGB images are synchronously acquired with tactile images. To address the reality gap of the visual modality, the following aspects are considered. 1) The real texture of the object is loaded. 2) Textures of the gripper, tactile sensors, and table are randomized within a specified range. 3) The camera's pose and resolution align with the physical parameters. In each trial, 50 images are rendered by randomly sampling the camera pose's degrees within a range of 2cm and 5°. 4) Gaussian noise is added to the images.

After lifting the object, the robot performs a shaking action to verify its stability, thus improving the robustness of the grasp outcomes. Finally, if the object is still in hand (i.e., the object's height along the z-axis $z > h$), the label $\ell = 1$. Otherwise, $\ell = 0$. The dataset is collected using 16 processes to obtain a total of $2.5 \times 10^6$ datapoints. The distribution of positive-to-negative labels in the dataset is around 6:4.

*C. Paired Tactile Dataset*

By using strictly paired simulation and real tactile images, one can achieve realistic optical properties and accurate contact dynamics mapping, thus enabling the effective transfer of tactile modality. However, the data collection policy needs to be carefully designed. To the best of our knowledge, no related work has been conducted on robotic grasping tasks. This work proposes a paired dataset collection policy for robotic grasping tasks, followed by image transfer using a modified conditional GAN.

The paired data collection procedure is shown in Fig. 3. Firstly, we select ten objects from the training object set. For each real object, we manually place a few stable poses. A real-time surface-based matching method implemented using the HALCON library [34] is used to estimate the object's 6D pose in each case. Secondly, the estimated pose is synchronized to the simulation environment via ROS service. In the simulation, 20 collision-free grasps corresponding to the synchronized pose are filtered from the grasp set containing 400 grasps. Finally, these grasps are synchronized sequentially to the real scene. Each synchronized grasp configuration involves conducting 20 grasping trials with the robot in each environment, starting with a gripping force of 10 N (the minimum gripping force of the gripper) and rising at an interval of 1 N. When closing the gripper, the paired images $(\mathcal{I}_r, \mathcal{I}_s)$ from real and simulation environments are recorded. The final dataset comprises 6800 paired data, which are partitioned into training (80%) and test (20%) sets.

*D. Training a Conditional GAN*

The migration network is built on the pix-to-pix GAN [26], as shown in Fig. 4. It aims to learn a mapping $f = G(x)$ from real tactile images to simulated images. Due to the utilization of simulated data for training the grasp stability prediction network, mapping real tactile images to simulated counterparts facilitates the direct deployment of the trained model on real hardware. The pix2pix architecture consists of a U-Net architecture as the generator $G$ and a patch-based fully convolutional network as the discriminator $D$. In our task, the objective of the discriminator $D$ is to reveal the differences between simulated and generated images, while the generator $G$ is to translate real images to simulated-like images to fool the discriminator $D$. We employ LSGAN loss [31] instead of original conditional GAN loss to promote training stability and generate higher-quality images. The discriminative loss and the generative loss are defined as follows,

$$\mathcal{L}_{LSGAN}(D) = \frac{1}{2}\mathbb{E}_{x,y}[(D(x,y)-1)^2]$$
$$+ \frac{1}{2}\mathbb{E}_x\left[D(x,G(x))^2\right] \quad (11)$$

$$\mathcal{L}_{LSGAN}(G) = \frac{1}{2}\mathbb{E}_x\left[(D(x,G(x))-1)^2\right] \quad (12)$$

Furthermore, we employ binary cross-entropy loss in Equation (8) to make the output of $G$ approach the simulated image as much as possible. Thus the final objective functions can be expressed as

$$\min\mathcal{L}(D) = \mathcal{L}_{LSGAN}(D) \quad (13)$$
$$\min\mathcal{L}(G) = \mathcal{L}_{LSGAN}(G) + \lambda\mathcal{L}_{BCE} \quad (14)$$

To optimize networks, we set $\lambda = 10$ and use the Adam optimizer with 2e-4 learning rate. We train the model with 10 batch sizes and 20 epochs.

## V. EXPERIMENTS AND DISCUSSION

Extensive experiments are conducted in simulation and real hardware to evaluate the proposed method. The experiments mainly include the following four aspects.

Firstly, comparative studies are conducted on a publicly available dataset and a self-collected dataset to compare the predictive performance of different models with different inputs. Secondly, we test the prediction accuracy of the multiple models in grasping trials. Thirdly, a delicate grasp experiment is designed to further verify the effectiveness of the proposed method. Finally, we use the SSIM metric to measure the effect of tactile image migration.

**TABLE I:** CROSS-VALIDATION RESULTS OF THE DIFFERENT MODELS ON PUBLIC DATASET.

|  | A (%) | P (%) | R (%) |
|---|---|---|---|
| Visual only | 62.2 ± 1.3 | 63.5 ± 0.6 | 61.1 ± 0.7 |
| Tactile only | 67.1 ± 0.4 | 66.2 ± 0.3 | 64.8 ± 0.5 |
| Calandra et al. [2] | 72.2 ± 0.6 | 72.8 ± 0.8 | 71.2 ± 0.6 |
| Ours-m | 76.6 ± 1.6 | 75.3 ± 1.1 | 75.1 ± 0.9 |
| **Ours** | **84.4 ± 0.7** | **85.2 ± 0.8** | **83.1 ± 0.5** |

**TABLE II:** CROSS-VALIDATION RESULTS OF THE DIFFERENT MODELS ON SELF-COLLECTED DATASET.

|  | A (%) | P (%) | R (%) |
|---|---|---|---|
| Visual only | 74.5 ± 0.8 | 76.1 ± 1.3 | 72.7 ± 0.6 |
| Tactile only | 83.5 ± 1.2 | 82.5 ± 0.5 | 81.5 ± 1.1 |
| Calandra et al. [2] | 90.6 ± 1.3 | 91.5 ± 1.5 | 88.3 ± 0.9 |
| Ours-m | 92.3 ± 0.7 | 94.3 ± 1.2 | 92.8 ± 0.5 |
| **Ours** | **98.3 ± 1.4** | **98.9 ± 0.8** | **97.2 ± 0.6** |

*A. Predictive Performance*

To comprehensively evaluate the proposed model, we conduct the experiments on a public dataset from Calandra *et al*. [2] and the dataset collected in Section IV with four baselines. The public dataset is collected by two GelSight sensors [33] and a Kinect camera mounted in front of the robot. This multimodal dataset contains 9269 grasping trials from 106 unique objects, and we use the data captured between closing gripper and lifting object as input to networks. `Ours` refers to the proposed attention-guided cross-modality fusion model. `Visual only` and `Tactile only` mask out the tactile input and visual input, respectively, and the pre-trained ResNet-50, along with the mentioned prediction head (Section III), is utilized for feature extraction and classification. The dual-stream network proposed by Calandra *et al.* [2] is adopted as the network architecture for direct visual-tactile fusion. Moreover, we denote `Ours-m` as a network structure in which only the last MSA is kept in the cross-modality fusion transformer, That is, the features from both channels are concatenated and passed through a self-attention module, as shown in Fig. 2. Finally, we employ the 3-fold cross-validation method to train the networks. Accuracy (A), Precision (P), and Recall (R) are chosen as the evaluation metrics.

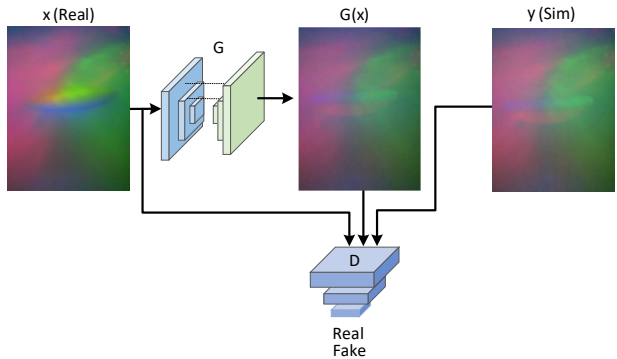

**Fig. 4:** Tactile migration network that maps real image to simulated image.

TABLE III: The Prediction Accuracy of the Grasping Experiment on Ten Objects, Presented in Percentage Form.

|  |  | obj. 1 | obj. 2 | obj. 3 | obj. 4 | obj. 5 | obj. 6 | obj. 7 | obj. 8 | obj. 9 | obj. 10 | Mean |
|---|---|---|---|---|---|---|---|---|---|---|---|---|
| Simulation | Visual only | 58 | 52 | 42 | 64 | 60 | 34 | 28 | 26 | 32 | 30 | 42.6 |
|  | Tactile only | 78 | 72 | 60 | 82 | 86 | 68 | 72 | 68 | 62 | 74 | 72.2 |
|  | Calandra et al. [2] | 90 | 84 | 80 | 96 | 100 | 90 | 82 | 86 | 70 | 78 | 85.6 |
|  | Ours | 100 | 92 | 96 | 100 | 100 | 100 | 90 | 94 | 84 | 100 | 95.6 |
| Real | Ours (WO-GAN) | 64 | 52 | 48 | 62 | 68 | 44 | 56 | 66 | 48 | 54 | 56.2 |
|  | Ours (W-GAN) | **86** | **78** | **70** | **92** | **96** | **80** | **82** | **84** | **70** | **94** | **83.2** |

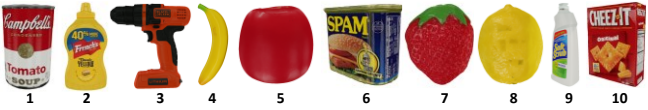

**Fig. 5:** In grasping trials, ten objects are utilized, with the first five selected from the training set and the final five from the test set.

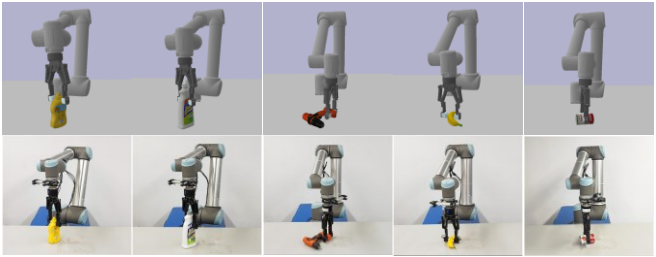

**Fig. 6:** A subset of the objects tested in simulated and real-world environments.

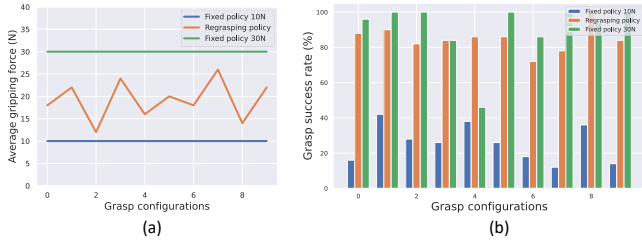

**Fig. 7:** Delicate grasping experimental results. (a) Average gripping force. (b) Grasp success rate is defined as the percentage of successful grasps to the total number of grasps.

The cross-validation results are reported in Table I and Table II. Overall, we see that `Visual only` method exhibits the lowest performance, whereas the the performance of `Tactile only` is second only to Calandra *et al.* [2]. This suggests that tactile feedback plays a more critical role than vision in predicting the grasp outcomes. Furthermore, the predictive power of the multimodal architectures is substantially improved compared to the unimodal ones. That is, integrating vision and tactile is indispensable for executing stable and gentle grasp operations. Additionally, the proposed full model `Ours` performs best, and the prediction performance has been dramatically improved compared to other baselines.

More specifically, we can see from Table I that the proposed method `Ours` exhibits an average performance improvement of around 12% and 8% over Calandra *et al.* [2] and `Ours-m`, respectively. And there is also a boost of around 8% and 5% in Table II. These findings demonstrate that our approach can effectively integrate visual and tactile modalities, leveraging their complementary strengths to improve the network's predictive power in a limited dataset. Furthermore, the comparison of Table I and Table II reveals that a large-scale dataset enables the network to undergo adequate training, significantly improving prediction performance.

### B. Grasping Performance

While testing on the dataset offers a preliminary assessment of the predictive power of different models, the goal is to evaluate the performance of the trained model in actual grasping trials. Therefore, we perform grasping tests on ten objects using both a simulation and an actual robot. Fig. 5 displays the ten objects used in our evaluation, comprising the first five from the training set and the remaining five from the test set. Fig. 6 shows the test scenarios for some objects in both simulation and real world. We conduct 50 grasping trials for each object with a randomized gripping force and employ prediction accuracy as a performance metric. This study uses GPG [28] to generate 6-DoF grasps from the single-view point cloud. GPG is a rapid solution that enables the sampling of parallel grasps from the 3D unknown point cloud. First, we evaluate four trained models in the simulation: `Visual only`, `Tactile only`, Calandra *et al.* [2], and `Ours`. Subsequently, we deploy our proposed model with and without the migration strategy on a real robot, which are denoted by `Ours (W-GAN)` and `Ours (WO-GAN)` in Table III, respectively. Before lifting, the grasp result is evaluated by the corresponding model.

The grasping results are shown in Table III. Upon analyzing the simulation results, we have observed that `Visual only` performs reasonably poorly. The main reason is that the gripping force changes during grasping trials, but the visual modality cannot perceive such slight variations. Comparatively, `Tactile only` shows relatively good prediction accuracy and generalization ability. This also again demonstrates the importance of tactile sense in delicate grasping tasks. The proposed model `Ours` achieves the highest accuracy and has strong generalization capability to unseen poses and objects, and the mean prediction accuracy attains a level of 95.6%. In actual tests, the mean accuracy of the proposed model employing the migration strategy is 83.2%, which is superior to the 56.2% accuracy attained via direct deployment. The findings further establish the efficacy of the migration strategy.

*C. Delicate Grasping*

The ultimate objective of evaluating grasping outcomes is to facilitate the generation of optimal grasping strategies, including minimum force grasping. This paper design a simplified rule to demonstrate the viability of the proposed model in delicate grasping experiments. We choose object 6 in Fig. 5 for our experiments on a real robot. Specifically, for a given grasp, let the robot start with a minimum force of 10N, increase by 1N each time, and the maximum force is 30N. During the process, the robot maintains a constant grasp pose and performs a lifting action when the model predicts a successful grasping result or the maximum gripping force is reached. Meanwhile, grasping with a fixed gripping force of 10N and 30N is the control group. We sample 10 grasp configurations, and 50 grasping trials are carried out for each configuration. The average value of the final gripping force and the grasp success rate are recorded.

We show the results in Fig. 7. For most grasping trials, the fixed policy with a 30N gripping force consistently yields the highest success rate, indicating that a higher gripping force generally leads to more stable grasps.

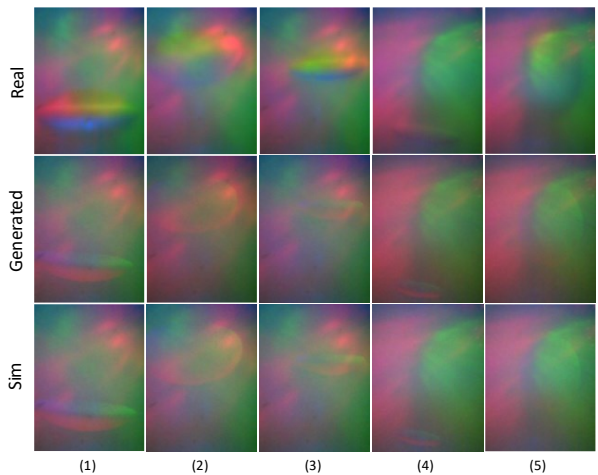

**Fig. 8:** Five sets of paired tactile images from the validation set. Each group consists of a real image acquired from the DIGIT sensor, an image generated by the conditional GAN, and a simulated image rendered by TACTO.

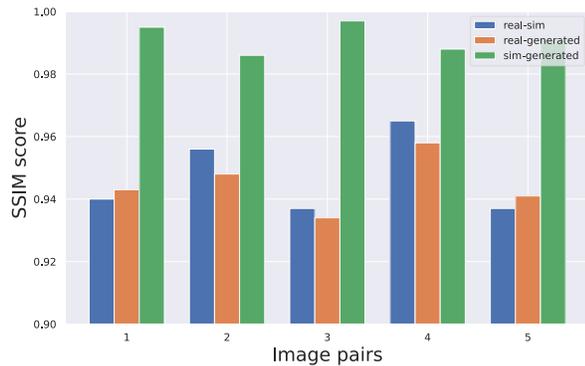

**Fig. 9:** The SSIM scores between real and simulated images, real and generated images, and simulated and generated images for each group in Fig. 8.

In contrast, our proposed model enables the robot to grasp objects with significantly less force, while still maintaining a similar success rate. Additionally, for unstable grasp configurations like grasp 5 in Fig. 7, the fixed policy using 30N results in a low success rate, whereas the regrasping policy based on grasp-stability evaluation achieves a relatively higher success rate.

*D. Tactile Modal Migration*

To evaluate the performance of the tactile transfer strategy visually, we utilize the trained generation model to produce images on the evaluation set, and employ the SSIM metric to quantify the similarity.

We compute SSIM scores between the generated and simulated tactile images on the validation set, resulting in an average score of 0.991, indicating a high degree of similarity. Additionally, we present five sets of paired tactile images in Fig. 8, with the corresponding SSIM scores displayed in Fig. 9.

## VI. CONCLUSION

This paper proposes an attention-guided cross-modality fusion network to assess grasp stability. This model is trained with synthetic visual and tactile images and then effectively deployed on a real robot using domain randomization and domain adaptation techniques. The experimental results show that our suggested model outperforms direct and co-attention fusion methods by approximately 12% and 8% on a publicly available small-scale dataset. Furthermore, the simulation and real-world grasping trials yield average prediction accuracies of 95.6% and 83.2%, respectively. The experimental findings demonstrate the effectiveness and efficiency of the proposed fusion method and transfer strategy in grasp stability evaluation tasks.

In future work, we will further investigate more effective transfer strategies for visual and tactile modalities. We also plan to introduce reinforcement learning to achieve minimum force grasping of unknown objects based on current work.

## ACKNOWLEDGMENT

The authors would like to express our most sincere gratitude to Zilin Si from Carnegie Mellon University and Shaoxiong Wang from Massachusetts Institute of Technology, who provided many valuable suggestions regarding tactile simulation.